\documentclass[11pt, a4paper]{article}

\usepackage{cite}
\usepackage{amsmath,amssymb,amsfonts}
\usepackage{algorithm,algpseudocode}
\usepackage{graphicx}
\usepackage{color}
\usepackage{varwidth}
\usepackage{epstopdf}

\begin{document}
\title{PSICA: decision trees for probabilistic subgroup identification with categorical treatments}

\author{Oleg Sysoev\thanks{
Department of Computer and Information Science ,
Link\"oping University,
58183, Link\"oping Sweden,
oleg.sysoev@liu.se}, Krzysztof Bartoszek\thanks{Department of Computer and Information Science ,
Link\"oping University,
58183, Link\"oping Sweden},
Eva-Charlotte Ekstr\"om\thanks{International Maternal and Child Health
Uppsala University,
75236, Uppsala Sweden},
Katarina Ekholm Selling\thanks{International Maternal and Child Health 
Uppsala University,
75236, Uppsala Sweden
}
}

\maketitle


\begin{abstract}
Personalized medicine aims at identifying best treatments for a patient with given characteristics. It has been shown in the literature that these methods can lead to great improvements in medicine compared to traditional methods prescribing the same treatment to all patients. Subgroup identification is a branch of personalized medicine which aims at finding subgroups of the patients with similar characteristics for which some of the investigated treatments have a better effect than the other treatments. A number of approaches based on decision trees has been proposed to identify such subgroups, but most of them focus on the two-arm trials (control/treatment) while a few methods consider quantitative treatments (defined by the dose). However, no subgroup identification method exists that can predict the best treatments in a scenario with a categorical set of treatments. We propose a novel method for subgroup identification in categorical treatment scenarios. This method outputs a decision tree showing the probabilities of a given treatment being the best for a given group of patients as well as labels showing the possible best treatments. The method is implemented in an R package \textbf{psica} available at CRAN. In addition to numerical simulations based on artificial data, we present an analysis of a community-based nutrition intervention trial that justifies the validity of our method.  
\end{abstract}

\textbf{keywords:} subgroup discovery; personalized medicine; decision trees; random forest; bootstrap

\section{Introduction}

It is very common that randomized trials are performed to investigate the efficiency of a new treatment. In these trials, a new treatment is compared to a control treatment, and if the new treatment is shown to be more efficient than the control it is suggested to be used on a population-wide level. Alternatively, in confirmatory subgroup analysis, effect of the treatment is investigated in the pre-specified subgroups \cite{mayer2015}.

Methods from personalized medicine \cite{jain2016} have drawn a lot of attention in medical and statistical literature \cite{hamburg2010}. These methods aim to identify and propose the best treatments to a patient with given characteristics (medical history). This clearly might lead to more efficient therapies than those proposed by confirmatory randomized trials. A lot of methods from personalized medicine are related to applications from genetics, i.e. these methods detect treatments that persons with specific genetic biomarkers benefit of. From a statistical perspective, this typically reduces to a high dimensional regression problem with binary input variables indicating the absence or the presence of corresponding genetic biomarkers. 

Another important category of personalized medicine is subgroup identification,  a comprehensive survey of methods from this category is provided in \cite{lipkovich2017}. The methods from this category identify subgroups of patients which benefit from the same treatments, and this identification can be based on the characteristics of various natures (binary, categorical, real-valued). Subgroup discovery methods can be applied to various experimental designs, including randomized clinical trials \cite{foster2011}.

Some subgroup identification methods are devoted to modeling optimal treatment regimes \cite{zhang2012,zhao2012,lu2013}. The primary purpose of these methods is to determine the optimal treatment for a given patient rather than detecting subgroups, but some efforts were done to summarize the outcome of such methods in a more descriptive way (e.g. \cite{foster2014}).  Due to a limited interpretability of results of optimal treatment regime methods, it can be hard to use them in general if policy making rather than black-box prediction of the optimal treatment is needed.

We focus on the subgroup identification methods that are inspired by the decision tree structures. Decision trees are easily interpretable which makes them very convenient for transparent policy making. A decision maker is thus not only able to see what treatments are recommended but also which patient characteristics this recommendation is based on. Some comparative analysis of such methods is presented in \cite{doove2014}.

It appears that the majority of subgroup identification methods focus on two-arm trials, i.e. when the treatment set is binary (control/treatment). Methods such as QUINT \cite{dusseldorp2014},  Virtual Twins (VT) \cite{foster2011}, Interaction Trees (IT) \cite{su2009}, SIDES \cite{lipkovich2011} are able to identify subgroups in the binary scenario. Being very efficient in some settings, all of these methods have peculiarities that in some situations can be considered as limitations. Most importantly, all these methods except of QUINT are focused on finding the groups when the treatment is better than the control but they ignore the situations when the inverse is true (called qualitative interaction). Among other peculiarities/limitations, one may mention inability of treatment of continuous effects (e.g. VT ), non-probabilistic nature of the algorithm (e.g. QUINT), overlapping subgroups (SIDES), providing an information about the mean difference in effect within a subgroup rather than stating the probability that one treatment is better than the other one (IT). A few methods go behind the binary scenario: recently, a method treating continuous treatments (ordered by dose) was proposed \cite{thomas2018}.

When the trials are performed with a categorical set of treatments, no methods exist that aim at finding subgroups and predicting which treatments are the best. In principle, Model-Based (MOB) trees \cite{zeileis2008} can be used to explain the dependence of the effect on the medical history variables (characteristics) and the treatment variable. However, because this method tries to explain the effect itself rather than the dominance of some treatments (prognostic variable problem, see \cite{loh2015}), very long trees might be needed to identify necessary subgroups. This makes conventional MOB trees very hard to use in practical policy making. It is also possible to apply the Gi method\cite{loh2015} to a scenario with a categorical set of treatments, but this method outputs mean effects per treatment and subgroup rather than specifying the best treatments. It means that when two or more treatments have the same effect, this method would not be able to identify such a situation due to randomness in the observed effect mean. In addition, this method produces trees that split ordinal predictors in the mean point, i.e. $x\leq \bar{x}$ and $x > \bar{x}$ which seriously limits the credibility of the resulting trees (unless they are very large which leads to poor interpretability).

We propose a novel method that is able to handle a scenario with categorical treatments (i.e. when two or more different types of treatments are considered). We call this method Probabilistic Subgroup Identification for CAtegorical treatments (PSICA). Our method is designed for randomized control trials and continuous effect variables. We believe that it of great importance for a subgroup identification method to provide statistical guarantees in the form of the probabilities of a treatment being the best for a given subgroup and, when data are not sufficient for a reliable decision, to state that there is no statistical guarantee that one of the treatments is more appropriate than the others. Accordingly, our method first uses random forests to compute the probabilities that a treatment is the best for a given patient, and then these probabilities are summarized by a decision tree in which each terminal node shows probabilities for a treatment to be the best and the label showing most likely treatments. When all probabilities are large enough within a node, its label may contain all treatments which is equivalent to saying  'I don't know which treatment is the best' (i.e. collect more data).

\begin{figure}[!t]
\centerline{\includegraphics[scale=.7]{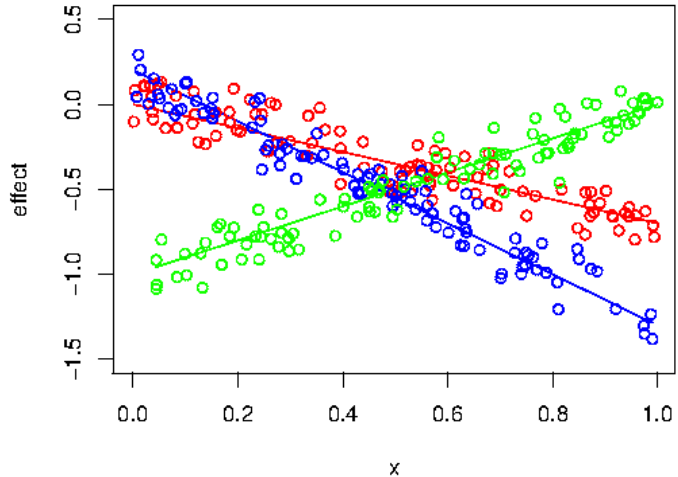}}
\caption{The effects for three treatments $\tau_1=A$, $\tau_2=B$ and $\tau_3=C$ are generated as $y(x,\tau_1)=-0.7 x+\epsilon$ (red),  $y(x,\tau_2)=-1.5 x+0.2+\epsilon$ (blue), $y(x,\tau_3)=x-1+\epsilon$ (green). Error term $\epsilon$ was generated as $N(0, 0.01)$}
\label{three}
\end{figure}

\begin{figure}[!t]
\centerline{\includegraphics[width=7cm]{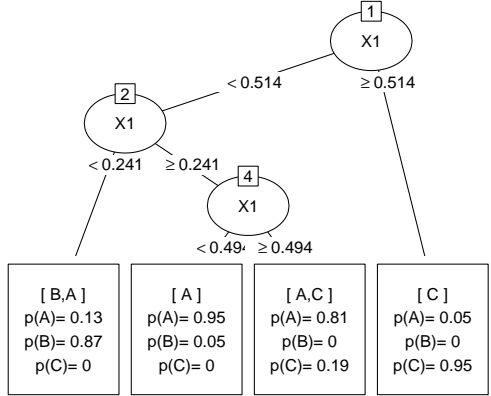}}
\caption{A PSICA tree showing subgroups, the probabilities of treatments being the best and labels containing the most likely optimal treatments.}
\label{threer}
\end{figure}

As an example, consider three treatments in which the effects are linear with respect to characteristics $x$. Figure \ref{three} demonstrates such example and some amount of observations corresponding to this setting. If the highest effect implies the best treatment, treatment B is supposed to be the best for smaller values of $x$, treatment $A$ should be the best for moderate $x$ and treatment C should be the best for the larger $x$. However, for smaller $x$ treatments A and B do not differ so much which means that a subgroup discovery method would probably have hard time to identify one best treatment. Figure \ref{threer} demonstrates the result of application of PSICA method to these data. It clearly illustrates that the subgroups are identified in the way that was expected.


In our numerical experiments, we compare PSICA with QUINT when there are two treatments. The QUINT method is chosen for comparisons because it is the only existing method capable of choosing the best treatment among two treatments and also stating when the treatments are equivalent. In addition, we use PSICA to perform subgroup identification for the MINIMat trial \cite{MINIMAT} that was conducted in Matlab sub-district, rural Bangladesh and contained 6 categorical interventions (treatments).

In section \ref{psicas}, we present PSICA method. In section \ref{numerical}, we present our numerical simulations and a real case study. Section \ref{concl} contains conclusions and discussions.

\section{PSICA trees}\label{psicas}

The problem of subgroup identification and some notations are introduced first. Given a data set $D=\left\{ (X_i, Y_i, t_i), i\in 1, \ldots, n \right\}$, where $X_i=(X_{i1}, \ldots, X_{ip})$ is a set of characteristics (input variables, predictors) for patient $i$, $Y_i$ is the effect of the given treatment $t_i$, where $t_i$ is one of the treatments that belong to the set $\mathcal{T}=\{\tau_1, \ldots, \tau_m\}$. We assume that $X_i$ values were obtained as a realization of a random variable $x$ with components $x^1, \ldots, x^p$. The response $Y_i$ is a realization of the effect $y$ for a given treatment $\tau$, and we assume that all treatments are possible to use for any patient.  The noise $\epsilon$ is assumed to be additive, i.e. $y(x,\tau)=f(x,\tau)+\epsilon$ where $f(x, \tau)$ is the expected effect for a given $x$ and $\tau$. The input variables may be categorical, ordinal or real-valued, and the effect is considered to be real-valued. 

In a binary setting, i.e. when $\mathcal{T}=(\tau_1, \tau_2)$, the subgroup identification problem can be defined as finding subgroups $G$ such that  and $$\pi(G,\tau_2, \tau_1)=p(Y(x,\tau_2)>Y(x,\tau_1)| x \in G)>1-\alpha$$ where $\alpha$ is some risk level, e.g. 0.05. This means that it is of interest to find subgroups of patients for which the second treatment is significantly better than the first one (which typically is a control treatment). Another interesting scenario is a qualitative subgroup identification, which means that the interesting subgroups are either those having $\pi(G,\tau_2,\tau_1)>1-\alpha$ or those satisfying $\pi(G,\tau_1,\tau_2)>1-\alpha$. 

When there are more than two treatments, the subgroup identification problem can be defined as follows:  identify groups $G$ and subsets of treatments $T \subset \mathcal{T}$ such that $$p(Y(x,\tau')>Y(x,\tau'')| x \in G,   \tau' \in T,  \tau'' \in \mathcal{T} \setminus T)>1-\alpha.$$ It means that we want either identify which treatments are useful and can be prescribed to a patient (treatments from $T$) or which treatments are useless for this subgroup and should not be given to these patients (treatments from $\mathcal{T} \setminus T$). Note that we require $T \neq \mathcal{T}$ because otherwise $\mathcal{T} \setminus T$ becomes empty and $(X,\tau'')$ becomes undefined.

PSICA trees partition the input space into non-overlapping regions and provide a label for each region and a probability distribution on the set $\{\tau_1, \ldots , \tau_m\}$ specifying how likely it is that a given treatment is the best one for the group of patients characterized by the input values from this region. The PSICA tree computation consist of two steps: estimation of distributions and growing the PSICA tree.

The first step of PSICA tree computation implies estimating distributions $\pi_k(x)$ which is a probability that the treatment $\tau_k$ is better than all alternative treatments for a given $x$. 
To estimate $\pi_k(x) , k=1,\ldots, m$ by simulation, we need to be able to generate samples from the joint distribution $p(y(x,\tau_1), \ldots, y(x, \tau_m))$. This distribution shows how likely it is that if a patient with characteristics $x$ receives treatment $\tau_1$ then the effect will be $y(x,\tau_1)$ and if the same patient receives $\tau_2$ then the effect will be $y(x,\tau_2)$ etc. If we are able to generate some number of samples $Y^b=(Y^b_1 (x), \ldots, Y^b_m (x)), b= 1, \ldots, B$ from this distribution, then $\pi_k(x)$ can be estimated as 

\begin{equation}
	\pi_k(x)=\frac{1}{B}\sum_{b=1}^B I(Y^b_k(x)> \max_{j=1,\ldots, m, j\neq k} Y^b_j(x)), \label{distr}
\end{equation}

To generate samples from $p(y(x,\tau_1), \ldots, y(x, \tau_m))$, we divide data $D$ into subsets $D_k=\left\{(X_i,Y_i,t_i): t_i=\tau_k, (X_i,Y_i,t_i) \in D\right\}$ for all $k=1,\ldots, m$. Each subset $D_k$ corresponds to one of the treatments. The partitioned data are further used to generate samples $Y^b$ by method 1 or method 2. 

Method 1 implies that $B$ data sets $D_k^b, b=1, \ldots, B$ are constructed by bootstrapping observations from $D_k$, we denote it as $D_k^b \sim Bootstrap(D_k)$. Afterwards, a machine learning model $M^b_k(x)$ is fit to each $D_k^b$ by using $y$ as response variable and $x$ as the set of predictor variables. We propose to use 
conditional inference random forest \cite{hothorn2006} models but in principle any other machine learning (regression) model can be employed. Finally, samples  $Y^b=(Y^b_1 (x), \ldots, Y^b_m (x))$ for any given $x$ are generated as $Y^b_k(x)=M^b_k(x), k=1, \ldots m, b=1, \ldots, B$.

Method 2 implies fitting a machine learning model to each $D_k$, estimating the prediction $M_k(x)$ and then estimating the variance $\sigma^2_k (x)$ of prediction by using the bias-corrected infinitesimal jackknife \cite{wager2014}. Finally, components $Y^b_k(x)$ of the samples are generated as $N(M_k(x), \sigma^2_k (x))$ for each $b=1, \ldots, B$. 

Estimation of $\pi_k(x)$ is summarized in Algorithm \ref{treatp}.

\begin{algorithm} 
\caption{Computation of the distributions of the treatment effects} \label{treatp}
\begin{algorithmic}

\State Given $D=\left\{ (X_i, Y_i, t_i), i\in 1, \ldots, n \right\}, t_i \in \mathcal{T}=\{\tau_1, \ldots, \tau_m\}$,  $method=1 \mbox{ or } 2$ .
 \For{$k=1$ {\bfseries to} $m$}
\State Compute $D_k=\{(X_i,Y_i,t_i): t_i=\tau_k, (X_i,Y_i,t_i) \in D,$ $ i=1,\ldots, n\}$
\EndFor

\If{$method=1$}
 \For{$b=1$ {\bfseries to} $B$, $k=1$ {\bfseries to} $m$ } 
\State $D^b_k \sim Bootstrap(D_k)$. 
\State Compute $M^b_k(x)$ from $D^b_k$ 
\State Compute $Y^b_k(X_i)=M^b_k(X_i)$ for each $(X_i, Y_i, t_i) \in D$
\EndFor
\EndIf

\If{$method=2$}
 \For{$k=1$ {\bfseries to} $m$}
\State Compute $M_k(x)$ and $\sigma^2_k (x)$ from $D_k$ 
\State \begin{varwidth}[t]{\linewidth} Generate $Y^b_k(X_i) \sim N(M_k(X_i), \sigma^2_k(X_i))$ for each $b=1, \ldots, B$, \par
		\hskip\algorithmicindent for each $(X_i, Y_i, t_i) \in D$
		\end{varwidth}
\EndFor
\EndIf
		
 \For{$i=1$ {\bfseries to} $n$, $k=1$ {\bfseries to} $m$ } 
\State Compute $\pi_k(X_i)$ by using (\ref{distr}) 
\EndFor
		

\end{algorithmic}
\end{algorithm}

The second step of PSICA tree computations implies growing a tree summarizing the probabilities $\pi_k(X_i)$ in such a way that interesting subgroups are discovered. We suggest two alternative methods for the tree growing process: method 3 and method 4. Method 3 requires growing a large tree and then letting a user to prune it until interpretable applied results are achieved and at the same time the tree becomes small enough to be used for policy making. Random forests are known to be robust to overfitting \cite{friedman2001} which means that for sufficiently large number of trees in the forests and for a sufficiently high number of bootstrap samples $B$ the probabilities reported in the tree nodes can be trusted. However, it can be hard to judge whether the settings used by user are sufficiently large which means that there might be a risk for producing spurious subgroups. To remedy this problem, we suggest method 4 which implies early pruning of the tree (prepruning) that guarantees that fewer spurious results are detected. However, since this method is based on hypothesis testing, there is a risk that some interesting subgroups are not found. 

Method 3 employs standard decision tree growing principles described in \cite{breiman1984}. More specifically, a data set $\Delta_0=\left\{(X_i, P_i), P_i=(\pi_1(X_i), \ldots, \pi_m(X_i) )\right\}$ with inputs $X_i$ and a vector response $P_i$ is constructed first. This data set is partitioned recursively by using various binary splitting rules $R_j$ (constructed differently for real-valued and categorical split variables) until some stopping criterion is met. This criterion might include constraints on the minimal amount of the observations in the node, maximal tree depth and other criteria. To decide which of splitting rules needs to be used, a data set $\Delta$ that corresponds to a tree node before split $R_j$ and data sets after this split $\Delta_1$ and $\Delta_2$ are  considered, and loss function values $L_1=L\left(\Delta\right)$, $L_2=L\left(\Delta_1\right)$ and $L_3=L\left(\Delta_2\right)$ are computed. A splitting rule that maximizes \textit{information gain}
\begin{equation}
g(\Delta, R_j)=L_1-(L_2+L_3) \label{g}
\end{equation}
is chosen to split the current node.

When no further split can be done, labels are assigned to the terminal nodes. In our settings, the following summary might be presented for a tree leaf corresponding to a data set $\Delta$:
\begin{itemize}
	\item Aggregated probabilities of each treatment being the best
	\begin{equation}
	\pi_k(\Delta)=\frac{1}{|\Delta|} \sum_{(X_i,P_i) \in \Delta} \pi_k(X_i), \label{aggr}
	\end{equation}
	where $|\Delta|$ denotes the number of observations in $\Delta$. 
	\item A set of \textit{useless treatments} $\mathcal{T}_u$ . The probabilities $\pi_k(\Delta)$ are sorted in increasing order as $\left(\pi_{k_1}(\Delta), \ldots, \pi_{k_m} (\Delta)\right)$ and $m'$ is found such that $\sum_{i=1}^{m'} \pi_{k_i}(\Delta) \leq \alpha$ and $\sum_{i=1}^{m'+1} \pi_{k_i}(\Delta) > \alpha$  where $\alpha$ is a risk level (e.g., $\alpha=0.05$). Set $\mathcal{T}_u$ is computed as $\mathcal{T}_u=\left\{\tau_{k_1}, \ldots, \tau_{k_{m'}}\right\}$ 
	\item A set of \textit{potential treatments} 
	\begin{equation}
			\mathcal{T}_p= \mathcal{T} \setminus \mathcal{T}_u \label{tp}
	\end{equation}
	
\end{itemize}

To enable successful subgroup identification, an appropriate loss function needs to be selected. To identify an appropriate function, it is important to consider how the resulting tree is going to be used in decision making. We assume that after a decision maker locates the patient into one of the terminal nodes of the decision tree, the aggregated probabilities $\pi_k(\Delta), k=1, \ldots, m$ are compared, and the treatments from $\mathcal{T}_u$ will be excluded by the decision maker. The remaining treatments are the potential treatments and ideally a further investigation of which of them should be prescribed to a given patient will be performed. However, it is also very likely that the aggregated probabilities corresponding to treatments from  $\mathcal{T}_p$ will be used by the decision maker directly as an indicator of which treatment should be used.

Therefore, we define the loss function $L(\Delta)$ as a cost of assignment of the individuals represented by $\Delta$ to the treatments that they do not benefit from. More specifically, we define \textit{truncated} probabilities as
\begin{equation}
\begin{split}
	\hat{\pi}_k (\Delta)= \frac{\pi_k (\Delta)}{\sum_{i \in \mathcal{T}_p}\pi_i (\Delta)}, k\in \mathcal{T}_p\\
	\hat{\pi}_k (\Delta)= 0, k \notin \mathcal{T}_p
	\end{split}
	\label{pihat}
\end{equation}
and therefore the cost of classifying a patient to a wrong treatment is 
\begin{equation}
\begin{split}
	\sum_{k=1}^m \sum_{j \in \mathcal{T}_p, j \neq k} c_{kj} p(\mbox{Assigned to }\tau_j  \mbox{ given } \tau_k \mbox{ is best})  \cdot p(\tau_k \mbox{ is best})
	\end{split}
	 \label{l}
\end{equation}
where $\left\{c_{kj}, k,j=1, \ldots, m\right\}$ are costs of giving the patient treatment $\tau_j$ given that his/her best treatment is $\tau_k$. A simple set of cost values is a zero-one cost: $c_{kj}=1$ when $k \neq j$ and zero otherwise. 

Equation (\ref{l}) can be rewritten as
\[
\sum_{k=1}^m \sum_{j \in \mathcal{T}_p, j \neq k} c_{kj} \hat{\pi}_j(\Delta) \cdot \pi_k(x)
\]

and, by summing up the loss values for the observations within $\Delta$, we obtain the following loss function: 

\begin{equation}
	L(\Delta)= \sum_{(X_i,P_i) \in \Delta} \sum_{k=1}^m \sum_{j \in \mathcal{T}_p, j \neq k} c_{kj} \hat{\pi}_j(\Delta) \cdot \pi_k(X_i) \label{Loss}
\end{equation}

If zero-one loss is used, it is easy to show that (\ref{Loss}) can be simplified as
\begin{equation}
L(\Delta)=|\Delta| \sum_{k=1}^m  \pi_k(\Delta) \cdot \left(1- \hat{\pi}_k(\Delta)\right) \label{loss}
\end{equation}

Method 4 involves early stopping to avoid discovery of spurious subgroups. The tree growing procedure is identical to the first approach described above with the only exception that the information gain function $g$ is modified in order to avoid splits that may generate spurious subgroups. More specifically, the modified information gain $g'$ is defined as $g'\left(\Delta, \Delta_1, \Delta_2\right)=g\left(\Delta, \Delta_1, \Delta_2\right)\cdot G\left(\Delta_1, \Delta_2\right)$ where $G\left(\Delta_1, \Delta_2\right)$ is equal to one if the distributions $\Pi_1 =\left\{\pi_k(\Delta_1), k=1, \ldots, m\right\}$ and $\Pi_2=\left\{\pi_k(\Delta_2), k=1, \ldots, m\right\}$ differ significantly and zero otherwise. 

To compute function $G$, we perform a chi-square test where we compare $\Pi_1$ and $\Pi_2$. For each $\Pi_j$, we compute counts 
\begin{equation}
\left\{n_{kj}=\left\lceil \pi_k(\Delta_j)\cdot |\Delta_j| \cdot \omega_j\right\rceil, k=1, \ldots, m\right\},
\label{chisq}
\end{equation}
where $\omega_j$ is an inflation factor defined by the standard deviation of the uniform distribution $U[0,1]$ divided by the standard deviation of $\left\{\pi_k(X_i): (X_i,P_i)\in \Delta_j\right\}$. The purpose of the inflation factor is to give higher weights to the distributions of $\pi_k(X_i)$ that have low variance (and, thus, more confident). After the counts for $\Pi_1$ and $\Pi_2$ are computed, these counts are combined into a two-way table, and the standard chi-square test is performed. If its p-value $p_{\chi}$ is lower than a risk level $\alpha$, we set $G=1$ otherwise $G=0$.

The summary of the PSICA tree growing strategy is given in Algorithm \ref{growprune}.

\begin{algorithm} 
\caption{Computation of PSICA tree} \label{growprune}
\begin{algorithmic}
\State Given $\Delta_0=\{(X_i, P_i): P_i=(\pi_1(X_i), \ldots, \pi_m(X_i) ),$ $i=1, \ldots, n \}$, $method= 3 \mbox{ or } 4$, risk level $\alpha$.
\State \textbf{Output}: $SplitNode(\Delta_0)$.

\Function{getLoss}{$\Delta$}
\State Compute $\pi_1 (\Delta), \ldots, \pi_m (\Delta)$ by using (\ref{aggr}).
\State Compute  $\hat{\pi}_1(\Delta), \ldots, \hat{\pi}_m(\Delta)$ by using (\ref{pihat})
\State Compute $L(\Delta)$ by using (\ref{loss})
\State \textbf{Output}: $L(\Delta)$
\EndFunction
\Function{getMask}{$\Delta_1, \Delta_2, \alpha$}
\State Compute $\left\{\pi_i (\Delta_j), i=1, \ldots, m, j=1,2\right\}$ by applying (\ref{aggr}).
\State Compute $n_{kj}$ by using (\ref{chisq}), $k=1, \ldots, m, j=1,2$
\State Compute $p$-value $p_{\chi}$  based on $\left\{n_{kj}, k=1, \ldots, m, j=1,2\right\}$.
\State Set $G=1$ if $p_{\chi}\leq \alpha$ and $G=0$ otherwise
\State \textbf{Output}: $G$
\EndFunction

\Function{Splitnode}{$\Delta$}
\If{Stopping criterion is met for $\Delta$}
\State \textbf{Output}: NULL
\Else

\For{$j=1$ {\bfseries to} $p$, each $R_j$}
	\State Split $\Delta= \Delta_1 \cup \Delta_2$ by using $R_j$. 
	\For{each $(\Delta_1, \Delta_2)$}
	\State Compute $g(\delta, R_j)= getLoss(\Delta)- getLoss(\Delta_1)-getLoss(\Delta_2)$
	\If{$method=4$}
		\State $g(\Delta, R_j) \leftarrow g(\Delta, R_j) \cdot getMask(\Delta_1, \Delta_2, \alpha)$
	\EndIf
	\EndFor 
\EndFor
\State Compute $R= \arg \max_{j,R_j}{g(\Delta, R_j)}$

\State {\bfseries Output: } $R$, $\Delta_1$, $\Delta_2$ , $Splitnode(\Delta_1)$ and $Splitnode(\Delta_2)$.
\EndIf
\EndFunction

\end{algorithmic}
\end{algorithm}

\section{Numerical experiments}\label{numerical}

Our PSICA method was implemented in an R package \textit{\textbf{psica}} which is available at CRAN \cite{psicaR}. To analyze the efficiency of the method, we tested it with the following models:

\begin{equation}
\begin{split}
	\mbox{M1 }:y(x, \tau)=\left(2\mbox{th} (2x)+3\right)I(\tau=\tau_1)+\\
	+\left(2\mbox{th} (x)+2.3\right) I(\tau=\tau_2)+\epsilon,
	\end{split}
\end{equation}

\noindent where $\epsilon \sim N(0,0.8^2)$ and $\mbox{th}(x)$ is a hyperbolic tangent function. Variance in this and the following models was adjusted in such a way that the highest signal-to-noise ratio is approximately 10. Some properties of this function considered at the interval $[-1,1]$ are that the function is relatively complex (i.e. includes nonlinearities) and that $\tau_1$ is best in the entire interval while the effect of $\tau_1$ and $\tau_2$ becomes very similar around $x=-0.5$. Therefore, one can expect that subgroup identification methods should be able to either to identify $\tau_1$ as the best treatment or they should be uncertain, for example around $x=-0.5$ and especially for smaller data sets. The QUINT method is aimed at finding qualitative interactions, i.e. it assumes that there exist regions where $\tau_1$ is better than $\tau_2$ and other regions where $\tau_2$ is better than $\tau_1$.  It means that this method is expected to fail in finding such interactions when the data are generated from M1. 

\begin{equation}
\begin{split}
	\mbox{M2/M3}:y(x, \tau)=0.5 I\left(x_1 \geq 0\mbox{ \& }x_2 \geq 0\right) I(\tau=\tau_1)+ \\
	+0.5 I\left(x_1<0\mbox{ \& }x_2<0\right) I(\tau=\tau_2)+\epsilon,
	\end{split}
\end{equation}
where $\epsilon \sim N(0, 0.2^2)$ (M2) and $\epsilon \sim Laplace(0, 0.2^2)$ (M3). These models contain qualitative interactions that are expected to be discovered by QUINT and also can be used to compare the effect of the error distribution (normal vs Laplace) on the performance of subgroup identification methods. 
\begin{equation}
\begin{split}
	\mbox{M4}:y(x, \tau)=\sum_{i=1}^{40} x_i + 5x_1 I(x_1 >0.5)I(\tau=\tau_1) + \\
	+5I(x_1 <0.5\mbox{ \& }x_2>0.5)I(\tau=\tau_2)+ \epsilon,
	\end{split}
\end{equation}
where $\epsilon \sim N(0,2^2)$. This model is interesting to analyze because it involves many variables in creating the main effect and a few variables that interact with the treatments. 
\begin{equation}
\begin{split}
	\mbox{M5 }:y(x, \tau)=\left(-0.7x-0.7\right)I(\tau=\tau_1)+\\
	+\left(-1.5x-1.1\right) I(\tau=\tau_2)+\\
	+\left(x-1\right) I(\tau=\tau_3)+\epsilon,
	\end{split}
\end{equation}
where $\epsilon \sim N(0,0.2^2)$. Model M5 is similar to the model explained in Figure \ref{three}. It contains 3 treatments and it can thus not be processed by the binary subgroup identification methods like QUINT. However, this model is good enough to study the behavior of PSICA model in a simple setting.
\begin{equation}
\begin{split}
	\mbox{M6 }:y(x, \tau)=\sum_{i=1}^{40} x_i + 5x_1 I(x_1 >0.5)I(\tau=\tau_1 \mbox{ or }\tau=\tau_2) +\\
	  +10(x_1 <0 \mbox{ \& }x_0='K1')I(\tau=\tau_3) + \epsilon,
	\end{split}
\end{equation}
where $\epsilon \sim N(0,2^2)$, and $\mathcal{T}=\{\tau_1, \ldots, \tau_4\}$. In this model, there is a main effect and also complex treatment effects: one subgroup benefits from treatments $\tau_1$ and $\tau_2$ while another subgroup benefits from treatment $\tau_3$. None of the patients benefit from $\tau_4$. This model also includes a categorical variable $x_0$ with 4 unique values, and this variable is important in defining one of the subgroups. This model can thus be regarded as good test of PSICA trees in real complex scenarios.

We perform the following numerical experiments 30 times. First, we generate data from models M1-M6 with $n$ observations where $n=300, 900$ or $1800$ and a randomized treatment assignment, where each $x$ components are generated as $U[-1,1]$. To make the analysis even more convincing, we add some number of irrelevant input variables generated as $U[-1,1]$ to each data set: 2 variables for models M1, M2, M3 and M5, 160 variables for M4 and M6. In the next step, we perform subgroup identification by using PSICA (for M1-M6) and QUINT (for M1-M4).  When computing PSICA trees, we use three alternatives: method $m_1$ denotes PSICA trees with probabilities computed by the bias-corrected infinitesimal jackknife (method=2 in Algorithm \ref{treatp}) and the number of variables per split it the random forest equal to the total amount of input variables $p$, method $m_2$ denotes PSICA trees with probabilities computed by the bias-corrected infinitesimal jackknife and the number of variables per split in the random forest equal to $\sqrt{p}$, method $m_3$ PSICA trees computed by the bootstrap approach (method=1 in Algorithm \ref{treatp}) and the number of variables per split in the random forest equal to $\sqrt{p}$.  Method $m_3$ is computed only for $n=300$ due to high computational burden. Other settings were specified as  $B=500$, $\alpha=0.05$, number of trees in a forest equal to 100, minimal amount of observations for splitting the node in a tree equal to $n/10$ in the trees belonging to forests and $n/5$ in the PSICA trees.  When computing QUINT trees ($m_4$), we use the bootstrap pruning and default settings specified in the corresponding R package \cite{dusseldorp2016}.

The trees obtained are analyzed by computing the following metrics: accuracy (a), uncertainty (u) and suspect (s). Given that for each feasible $x$ a tree delivers the predicted best treatments $\mathcal{T}_p$ while the true best treatments are $T_p$,  metrics $a$ and $u$ are defined as follows:
\begin{equation}
	a(D)=\frac{1}{n}\sum_{(X,Y,T) \in D} I(T_p(X) \subseteq \mathcal{T}_p(X) )
\end{equation}
\begin{equation}
	u(D)=\frac{1}{n} \sum_{(X,Y,T) \in D} I\left(|\mathcal{T}_p(X)| > |T_p(X)| \right)
\end{equation}

where $|S|$ denotes the number of elements in a set $S$. Accuracy values represent proportions of the correct predictions while the uncertainty values specify how  uncertain the tree is. Note that a tree can in principle achieve $100\%$ accuracy by predicting all treatments as a full set $\mathcal{T}$, but it will also imply $100\%$ uncertainty. 

The suspect $s(\Delta)$ is defined  as a the sum of the amounts of observations corresponding to the nodes that are immediately above the irrelevant splits divided by the sum of the amounts of observations corresponding to all nodes in the tree. Therefore, if an irrelevant variable is located in the top levels of the tree, the suspect value is expected to be high.

For PSICA trees, we also compute a measure which we call \textit{decision accuracy}. As it was discussed in section \ref{psicas}, we assume that when a PSICA tree returns a label with more than one treatment, a decision maker is ideally supposed to make a further investigation regarding which of these treatments should be given to a patient. However, it is also likely that the decision maker will use the aggregated probabilities shown in the corresponding tree node to make a decision. However this might not be a good strategy in some situations. Suppose $\mathcal{T}=\{\tau_1, \tau_2\}$ and in the given tree node $\pi_1=0.45$ and $\pi_2=0.55$. Although treatment $\tau_2$ has a somewhat higher probability, it is clear that the model is quite unsure about which treatment is the best one for the group of patients associated with the given tree node. This means that in this case a further investigation is probably the most reasonable option. Assume though that the PSICA tree returns a set of truncated probabilities $\left\{\hat{\pi}_k(x), k=1, \ldots, m\right\}$ for a given $x$ and the decision maker makes a decision as $\tilde{\tau}(x) \sim Multinomial(\hat{\pi}_1(x), \ldots, \hat{\pi}_m(x))$. Decision accuracy measures the proportion of the correct decisions in this scenario as
\begin{equation}
	\delta(D)=\frac{1}{n} \sum_{(X,Y,T) \in D} I\left( \tilde{\tau}(X) \in T_p(X) \right)
\end{equation}

\begin{table}[!t]
  \centering
  \caption{Mean accuracy rates (over 30 experiments) for different data models (M1-M6) processed by four methods ($m_1-m_4$). Standard error of the mean is specified in parentheses.}
    \begin{tabular}{rrrrllr}
    \multicolumn{1}{l}{n} & \multicolumn{1}{l}{model} & \multicolumn{1}{l}{$m_1$} & \multicolumn{1}{l}{$m_2$} & $m_3$   & $m_4$    & \multicolumn{1}{l}{\% fails in $m_4$} \\
300   & 1     & 1.00 (0.000) & 1.00 (0.000) & 1.00 (0.000) & 0.83 (0.009) \\
   300   & 1     & 1.00 (0.000) & 1.00 (0.000) & 1.00 (0.000) & 0.83 (0.009) & 0.90 \\
    900   & 1     & 1.00 (0.000) & 1.00 (0.000) & --    & --    & 1.00 \\
    1800  & 1     & 1.00 (0.000) & 1.00 (0.000) & --    & --    & 1.00 \\
    300   & 2     & 0.99 (0.004) & 1.00 (0.000) & 0.96 (0.005) & 0.80 (0.033) & 0.40 \\
    900   & 2     & 0.99 (0.002) & 1.00 (0.000) & --    & 0.94 (0.011) & 0.40 \\
    1800  & 2     & 1.00 (0.001) & 1.00 (0.000) & --    & 0.96 (0.007) & 0.33 \\
    300   & 3     & 0.99 (0.002) & 1.00 (0.000) & 0.95 (0.007) & 0.71 (0.044) & 0.26 \\
    900   & 3     & 0.99 (0.001) & 1.00 (0.000) & --    & 0.94 (0.009) & 0.33 \\
    1800  & 3     & 1.00 (0.001) & 1.00 (0.000) & --    & 0.96 (0.007) & 0.43 \\
    300   & 4     & 1.00 (0.001) & 1.00 (0.000) & 0.99 (0.003) & 0.59 (0.026) & 0.33 \\
    900   & 4     & 0.99 (0.002) & 1.00 (0.000) & --    & 0.99 (0.001) & 0.70 \\
    1800  & 4     & 0.99 (0.001) & 1.00 (0.000) & --    & 0.99 (0.001) & 0.86 \\
    300   & 5     & 0.97 (0.004) & 1.00 (0.002) & 0.97 (0.004) & --    & 1.00 \\
    900   & 5     & 0.97 (0.004) & 1.00 (0.001) & --    & --    & 1.00 \\
    1800  & 5     & 0.97 (0.005) & 1.00 (0.000) & --    & --    & 1.00 \\
    300   & 6     & 0.87 (0.028) & 0.88 (0.041) & 0.64 (0.048) & --    & 1.00 \\
    900   & 6     & 0.89 (0.007) & 0.89 (0.039) & --    & --    & 1.00 \\
    1800  & 6     & 0.87 (0.004) & 0.86 (0.039) & --    & --    & 1.00 \\

    \end{tabular}%
		  \label{acc}%
\end{table}%

\begin{table}[!t]
  \centering
  \caption{Mean uncertainty rates (over 30 experiments) for different data models (M1-M6) processed by four methods ($m_1-m_4$). Standard error of the mean is specified in parentheses.}
    \begin{tabular}{rrrrll}
    \multicolumn{1}{l}{n} & \multicolumn{1}{l}{Model} & \multicolumn{1}{l}{$m_1$} & \multicolumn{1}{l}{$m_2$} & $m_3$ & $m_4$ \\

    300   & 1     & 0.31 (0.018) & 0.44 (0.036) & 0.18 (0.025) & 0.44 (0.031) \\
    900   & 1     & 0.26 (0.019) & 0.55 (0.041) & --    & -- \\
    1800  & 1     & 0.22 (0.018) & 0.73 (0.040) & --    & -- \\
    300   & 2     & 0.24 (0.022) & 0.50 (0.008) & 0.11 (0.022) & 0.39 (0.034) \\
    900   & 2     & 0.17 (0.017) & 0.50 (0.004) & --    & 0.52 (0.018) \\
    1800  & 2     & 0.21 (0.016) & 0.50 (0.002) & --    & 0.51 (0.015) \\
    300   & 3     & 0.30 (0.026) & 0.50 (0.005) & 0.10 (0.019) & 0.32 (0.050) \\
    900   & 3     & 0.21 (0.015) & 0.50 (0.005) & --    & 0.52 (0.017) \\
    1800  & 3     & 0.26 (0.019) & 0.50 (0.002) & --    & 0.53 (0.016) \\
    300   & 4     & 0.41 (0.015) & 0.44 (0.004) & 0.40 (0.015) & 0.33 (0.036) \\
    900   & 4     & 0.23 (0.025) & 0.43 (0.003) & --    & 0.56 (0.004) \\
    1800  & 4     & 0.08 (0.011) & 0.44 (0.002) & --    & 0.57 (0.003) \\
    300   & 5     & 0.36 (0.026) & 0.68 (0.029) & 0.35 (0.031) & -- \\
    900   & 5     & 0.33 (0.024) & 0.71 (0.024) & --    & -- \\
    1800  & 5     & 0.26 (0.020) & 0.85 (0.027) & --    & -- \\
    300   & 6     & 0.41 (0.007) & 0.42 (0.004) & 0.40 (0.008) & -- \\
    900   & 6     & 0.15 (0.018) & 0.41 (0.003) & --    & -- \\
    1800  & 6     & 0.06 (0.005) & 0.42 (0.002) & --    & -- \\

    \end{tabular}%
  \label{unc}%
\end{table}%

\begin{table}[!t]
  \centering
  \caption{Mean suspect rates (over 30 experiments) for different data models (M1-M6) processed by four methods ($m_1-m_4$). Standard error of the mean is specified in parentheses.}
    \begin{tabular}{rrrrll}
		\multicolumn{1}{l}{n} & \multicolumn{1}{l}{Model} & \multicolumn{1}{l}{$m_1$} & \multicolumn{1}{l}{$m_2$} & $m_3$ & $m_4$ \\
300   & 1     & 0.08 (0.030) & 0.09 (0.041) & 0.15 (0.051) & 0.11 (0.033) \\
    900   & 1     & 0.05 (0.017) & 0.01 (0.010) & --    & -- \\
    1800  & 1     & 0.04 (0.014) & 0.04 (0.014) & --    & -- \\
    300   & 2     & 0.00 (0.000) & 0.00 (0.000) & 0.01 (0.007) & 0.02 (0.011) \\
    900   & 2     & 0.00 (0.000) & 0.00 (0.000) & --    & 0.02 (0.007) \\
    1800  & 2     & 0.00 (0.000) & 0.00 (0.003) & --    & 0.00 (0.004) \\
    300   & 3     & 0.00 (0.000) & 0.00 (0.003) & 0.00 (0.003) & 0.06 (0.016) \\
    900   & 3     & 0.00 (0.000) & 0.01 (0.005) & --    & 0.03 (0.008) \\
    1800  & 3     & 0.00 (0.000) & 0.00 (0.000) & --    & 0.01 (0.006) \\
    300   & 4     & 0.01 (0.008) & 0.00 (0.000) & 0.02 (0.014) & 0.48 (0.054) \\
    900   & 4     & 0.00 (0.000) & 0.00 (0.000) & --    & 0.00 (0.000) \\
    1800  & 4     & 0.00 (0.000) & 0.00 (0.000) & --    & 0.00 (0.000) \\
    300   & 5     & 0.00 (0.000) & 0.01 (0.007) & 0.02 (0.009) & -- \\
    900   & 5     & 0.00 (0.000) & 0.00 (0.000) & --    & -- \\
    1800  & 5     & 0.00 (0.000) & 0.01 (0.007) & --    & -- \\
    300   & 6     & 0.06 (0.026) & 0.00 (0.000) & 0.04 (0.018) & -- \\
    900   & 6     & 0.01 (0.006) & 0.00 (0.000) & --    & -- \\
    1800  & 6     & 0.00 (0.000) & 0.00 (0.000) & --    & -- \\

    \end{tabular}%
  \label{sus}%
\end{table}%

\begin{table}[!t]
  \centering
  \caption{Mean decision accuracy rates (over 30 experiments) for different data models (M1-M6) processed by PSICA methods ($m_1-m_3$). Standard error of the mean is specified in parentheses.}
    \begin{tabular}{rrrrl}
		\multicolumn{1}{l}{n} & \multicolumn{1}{l}{Model} & \multicolumn{1}{l}{$m_1$} & \multicolumn{1}{l}{$m_2$} & $m_3$ \\

    300   & 1     & 0.90 (0.007) & 0.89 (0.009) & 0.94 (0.009) \\
    900   & 1     & 0.93 (0.007) & 0.87 (0.009) & -- \\
    1800  & 1     & 0.95 (0.005) & 0.84 (0.007) & -- \\
    300   & 2     & 0.96 (0.002) & 0.88 (0.007) & 0.97 (0.004) \\
    900   & 2     & 0.97 (0.002) & 0.89 (0.005) & -- \\
    1800  & 2     & 0.97 (0.002) & 0.88 (0.004) & -- \\
    300   & 3     & 0.95 (0.004) & 0.89 (0.007) & 0.98 (0.003) \\
    900   & 3     & 0.97 (0.002) & 0.88 (0.005) & -- \\
    1800  & 3     & 0.96 (0.002) & 0.87 (0.004) & -- \\
    300   & 4     & 0.89 (0.009) & 0.80 (0.004) & 0.88 (0.009) \\
    900   & 4     & 0.95 (0.003) & 0.83 (0.002) & -- \\
    1800  & 4     & 0.97 (0.002) & 0.84 (0.002) & -- \\
    300   & 5     & 0.88 (0.008) & 0.76 (0.010) & 0.87 (0.009) \\
    900   & 5     & 0.91 (0.005) & 0.75 (0.005) & -- \\
    1800  & 5     & 0.91 (0.004) & 0.71 (0.006) & -- \\
    300   & 6     & 0.87 (0.009) & 0.79 (0.005) & 0.86 (0.008) \\
    900   & 6     & 0.95 (0.004) & 0.83 (0.003) & -- \\
    1800  & 6     & 0.95 (0.003) & 0.84 (0.002) & -- \\
    \end{tabular}%
  \label{d}%
\end{table}%

Tables \ref{acc}, \ref{unc},  \ref{sus} and \ref{d} demonstrate the results of our simulation experiments. It can be concluded that $m_1$, $m_2$ and $m_3$ provide a similar accuracy which is close to $100\%$ in the majority of scenarios, both when binary treatments are used and when categorical scenarios are considered. Method $m_4$ (QUINT) has lower accuracy values, especially for smaller data. It can also be observed that $m_4$ quite often fails to find any qualitative interaction when they exist (models M2-M4). For a more complex model (M4) this failure rate goes up to 80\%. By comparing the accuracies of methods $m_1$-$m_3$ across M2 and M3, no noticeable difference can be detected which indicates that PSICA trees are not so sensitive to the error distribution.

When comparing uncertainties, an interesting fact can be observed: allowing the conditional inference random forest to use all input variables at any split leads to lower uncertainty rates than the setting $\sqrt{p}$ variables at any split recommended in the literature for the random forests. This happens because the trees in the forests are grown by means of early stopping involving hypothesis testing: if there are no relevant variables in the randomly selected subset of input variables, the corresponding tree node will not be split further. It may lead to deficient trees in some cases. It can also be observed that uncertainty rates decrease with increasing sample size for model $m_1$, while for models $m_2$ and $m_4$ these rates usually do not change much or sometimes increase. Noticeably, uncertainty rates of $m_1$ are generally lower than those of $m_2$ and $m_4$, and the uncertainty rates of $m_2$ are generally comparable to the rates of $m_4$ with the exception of a more complex model (M4) where the $m_2$ rates are lower than the $m_4$ rates. The uncertainty rates of $m_3$ are lower that those of $m_1$ for all models that were available for comparisons indicating than applying bootstrap instead of the variance approximation might lead to better decisions. The price of this is a much higher computational time.   Table \ref{sus} illustrates that both QUINT and PSICA are good in finding relevant predictors: the suspect rates change approximately between 0 and 0.1 for both methods. However, when a complex model is considered and the data are small, $m_4$ appears to have problems in finding the relevant predictors since the suspect rate becomes 0.48 in this case.

Table \ref{d} demonstrates that the decision accuracies for method $m_1$ is often above $0.87$ (i.e. $87\%$) and they grow with the increasing sample size. Method $m_2$ has somewhat lower decision accuracies which confirms our previous finding: using all input variables at any split leads to a better performance of PSICA trees. Results generated by method $m_3$ again indicate that using second-level bootstrap might lead to more accurate results compared to the bias-corrected infinitesimal jackknife.

In addition to the simulation experiments, we analyze the so called MINIMat data \cite{MINIMAT} with the PSICA method. The MINIMat trial was conducted in the Matlab sub-district, rural Bangladesh. In this area, 4436 pregnant women were enrolled between November 2001 and October 2003 to take part in the trial. The design and interventions of the MINIMat trial has previously been described in detail \cite{persson2012}. Very briefly, pregnant women were individually and randomly allocated in a 2 by 3 factorial design into two prenatal food- and three micronutrient supplementation groups. Food supplementation was promoted to start either in early pregnancy (E for early) or at the women's own liking (U for usual). The three micronutrient groups were: 30~mg of iron supplementation (X), 60~mg iron (Y), and 30~mg of iron, 400~mg of folic acid, and 13 other micronutrients (Z). At enrollment and during pregnancy, characteristics of the women and their households were collected. In this example, 124 variables, such as maternal anthropometry, parity, education, morbidity, exposure to domestic violence, as well as household food insecurity and assets, during the time of pregnancy were included as inputs. 
\begin{figure}[!t]
\centerline{\includegraphics[width=15cm]{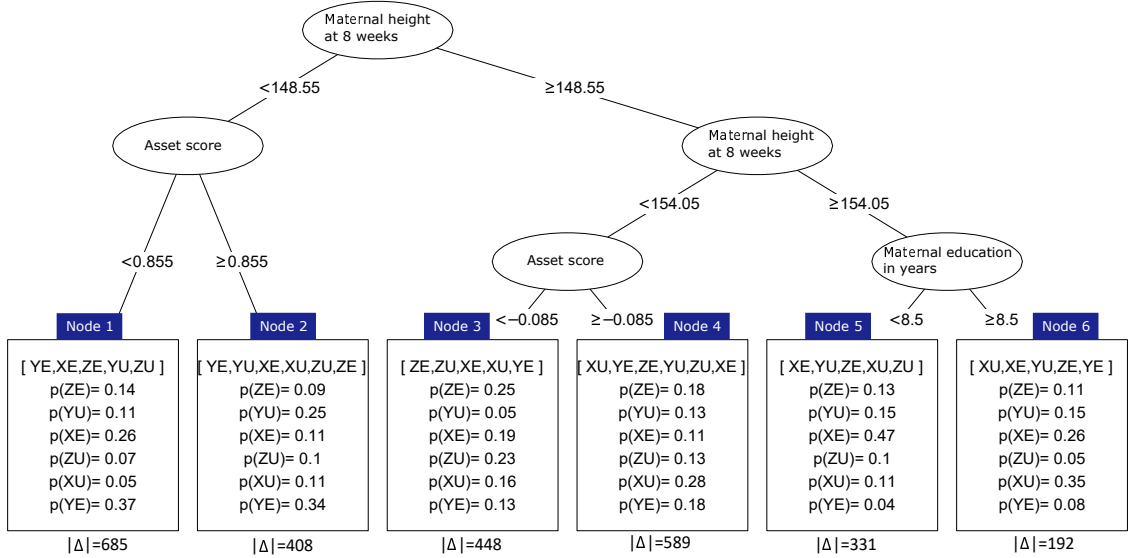}}
\caption{A PSICA tree showing subgroups and probabilities of various treatments for the MINIMat trial. Amounts of observations in the terminal nodes are represented by $|\Delta|$}
\label{minimat}
\end{figure}

The effect variable is the children's height-for-age z-score at 54 months (HAZ), and the aim of PSICA tree analysis is to find out which interventions increase HAZ of the children. We computed a PSICA tree with pruning, $B=1000$, $\alpha=0.05$, number of trees in a forest equal to 100, minimal amount of observations for splitting the node in a tree equal to $40$ in the trees belonging to forests and equal to $60$ in the PSICA tree.

Figure  \ref{minimat} shows that in 4 out 6 nodes (Nodes 1, 2, 3, 5) supplementation options including early food supplementation had a larger probability of increasing HAZ at 54~m than the usual food supplementation, and this is in agreement with previous results of the trial, see \cite{khan2011}, Table 3. Similarly our finding that in 3 out 6 nodes (Nodes 4, 5 and 6) a supplementation including 30~mg of iron, and in 2 out 6 nodes (Nodes 1 and 2) supplementation containing 60~mg of iron had higher HAZ compared to multiple micronutrient supplementation is also in agreement with previous results, see \cite{khan2011}, Table 3.

A result that has not been shown previously is that the optimal micronutrient supplementation varied with maternal height. Among the shortest women (Nodes 1 and 2) supplements containing 60~mg had the highest probability of a better HAZ in their offsprings. Among taller women (Nodes 3-6) supplements with a lower amount of iron (30~mg and MMS) had higher probabilities, and in 3 out these 4 nodes the optimal supplementation was 30~mg. 
While these differences in effects on young child height development have not previously been shown they are biologically plausible in that shorter women are likely to have experienced more of nutrients deficiencies and thus larger nutrient requirements such as a larger dose of iron may be needed for optimal growth of their children. Maternal height has been shown to modify effect of micronutrient supplementation on other early life outcomes \cite{smith2017} and it is reasonable to believe that it will also modify other later outcomes.  Similarly indicators of socio-economic situation such as maternal education has been shown to modify effect of micronutrient supplementation on early life outcome \cite{smith2017} and thus may also be of importance for young child height. The importance of iron for fetal, infant and child growth have been shown in studies in low-income settings \cite{nisar2016,nguyen2017} and iron supplementation has been highlighted as a key intervention to improve maternal and child health \cite{bhutta2008}.

\section{Conclusions and discussion} \label{concl}

In this work, we introduce PSICA trees which is a novel method for subgroup identification in scenarios with categorical sets of treatments. Our numerical results illustrate that, with appropriate settings, PSICA trees provide high accuracies of prediction of the best treatments and the method's uncertainty decreases with an increasing amount of data. At the same time, PSICA trees are easily interpretable and can therefore be used for policy making. PSICA trees seem to be able to identify meaningful subgroups even when there is a moderate mean effect from a lot of inputs, while in these cases the QUINT method often fails to identify meaningful subgroups or it gives low accuracies. PSICA trees are also able to handle cases when none of the treatments has a significantly better effect than the other treatment: in this case a non-informative tree (i.e. in which all the treatments are declared to be best) can be returned.

It appears that PSICA trees providing the best accuracies are obtained when the amount of splitting variables in the corresponding random forest is equal to the total amount of inputs. There is also an indication of that bootstrapping random forests instead of using bias-corrected infinitesimal jackknife might lead to lower uncertainties of the PSICA method. However, the price for this is a greatly increased computational time. Some of the results also indicate that PSICA trees are not very sensitive to error distribution.

PSICA trees are computed by estimating probabilities and loss functions in a statistically motivated manner which leads to high accuracies and low suspect rates in our simulation experiments. Real case studies justify the validity of our method because the information provided by PSICA tree is also confirmed by other medical studies. 

PSICA trees presented in this paper have some limitations. First, the PSICA method was described for real-valued effect variables only. In addition, it is assumed that randomized clinical trials data is used. Accordingly, a further research direction is to generalize the PSICA algorithm to  categorical effect scenarios and to investigate how it needs to be modified for non-randomized trials. In addition, investigating possibilities of post-pruning instead of pre-pruning might decrease uncertainty rates of the PSICA method.

\section{Acknowledgment}
Authors are grateful to the Swedish Research Council for providing funding for this work (project 2014-2161). We would also like to thank A. Rahman, S. el Arifeen, R. Naved, A. I. Khan and L. \AA. Persson for collecting and providing MINIMat data for our analysis.

\bibliographystyle{plain}
\bibliography{psica}

\end{document}